\documentclass[11pt,draftclsnofoot,onecolumn]{IEEEtran}

\usepackage{times}
\usepackage{epsfig}
\usepackage{graphicx}
\usepackage{amsmath}
\usepackage{amssymb}
\usepackage{overpic}

\usepackage{algorithm}
\usepackage{algorithmic}

\usepackage{url}
\usepackage{color}

\def\hrho{g^}

\def\nz{{ v}}
\def\nx{{ z}}

\def\ffun{\nu}
\def\gfun{u}

\def\A{{\mathbf A}}
\def\B{{\mathbf B}}

\def\D{{\mathbf D}}

\def\HH{{\mathbf H}}
\def\II{{\mathbf I}}

\def\PP{{\mathbf P}}

\def\ii{{\mathbf e}}

\def\g{{\bf \mathfrak{g}}}
\def\kk{{\mathbf k}}

\def\n{{\mathbf n}}

\def\w{{\mathbf w}}
\def\wbar{\bar{{\mathbf w}}}
\def\x{{\mathbf x}}
\def\y{{\mathbf y}}

\def\brho{{\boldsymbol \rho}}

\newtheorem{Thm}{Theorem}
\newtheorem{Def}{Definition}

\def\GM{{\boldsymbol \Gamma}}
\def\gam{{\boldsymbol \gamma}}

\newcommand{\myendofproof}[0]{\hfill $\blacksquare$ \newline}

\usepackage{graphicx}
\usepackage{subfigure}
\usepackage{overpic}

\graphicspath{{Figures/}}

\title{Non-Uniform Blind Deblurring with a Spatially-Adaptive Sparse Prior}

\author{Haichao Zhang\thanks{School of Computer Science, Northwestern Polytechnical University, Xi'an, P.R. China ({\tt hczhang1@gmail.com}).}
        and  David Wipf\thanks{Microsoft Research Asia,
        Beijing, P.R. China ({\tt davidwipf@gmail.com}).}}

\begin{document}
\maketitle

\begin{abstract}
Typical blur from camera shake often deviates from the standard uniform convolutional script, in part because of problematic rotations which create greater blurring away from some unknown center point.  Consequently, successful blind deconvolution requires the estimation of a spatially-varying or non-uniform blur operator.  Using ideas from Bayesian inference and convex analysis, this paper derives a non-uniform blind deblurring algorithm with several desirable, yet previously-unexplored attributes.  The underlying objective function includes a spatially adaptive penalty which couples the latent sharp image, non-uniform blur operator, and noise level together.  This coupling allows the penalty to automatically adjust its shape based on the estimated degree of local blur and image structure such that regions with large blur or few prominent edges are discounted.  Remaining regions with modest blur and revealing edges therefore dominate the overall estimation process without explicitly incorporating structure-selection heuristics.  The algorithm can be implemented using a majorization-minimization strategy  that is virtually parameter free.  Detailed theoretical analysis and empirical validation on real images serve to validate the proposed method.
\end{abstract}

\section{Introduction}\label{sec:Intro}
Image blur is an undesirable degradation that often accompanies the image formation
process and may arise, for example, because of camera shake during acquisition. Blind image deblurring  strategies aim to recover a sharp image from only a blurry, compromised observation.  Extensive efforts have been devoted to the uniform blur (shift-invariant) case, which can be described  with the convolutional observation model
\begin{equation}
\y = \kk\ast \x + \n,
\end{equation}
where $\ast$ denotes 2D convolution, $\x$ is the unknown sharp image, $\y$ is the observed blurry image, $\kk$ is the unknown blur kernel (or point spread function), and
$\n$ is a zero-mean Gaussian noise term with covariance $\lambda\II$ ~\cite{Fergus06removingcamera,hqdeblurring_siggraph2008,LevinWDF11_PAMI, fast_motion_deblur_2009, XuJ10_ECCV, norm_sparse,WangYYZ08,BeckT09}.  Unfortunately, many real-world photographs contain blur effects that vary across the image plane, such as when unknown rotations are introduced by camera shake~\cite{LevinWDF11_PAMI}.

More recently, algorithms have been generalized to explicitly handle some degree of non-uniform blur using the more general observation model
\begin{equation}
\y = \HH \x + \n,
\end{equation}
where now (with some abuse of notation) $\x$ and $\y$ represent vectorized sharp and blurry images respectively and each column of the blur operator $\HH$ contains the spatially-varying effective blur kernel at the corresponding pixel site ~\cite{Whyte_non-uniformdeblurring, Gupta10singleimage, HarmelingHS_NIPS10, HirschSHS_ICCV11, hu_bmvc2012, ChoCTL12, Xu_depth-awaremotion, non_uniform_restoration_chp3, JiHui12}.  Note that the original uniform blur model can be achieved equivalently when $\HH$ is forced to adopt a simple toeplitz structure.  In general, non-uniform blur may arise under several different contexts.  This paper will focus on the blind removal of non-uniform blur caused by general camera shake (as opposed to blur from object motion) using only a single image, with no additional hardware assistance.

\begin{figure}[t]
\centering
\includegraphics[height=6cm]{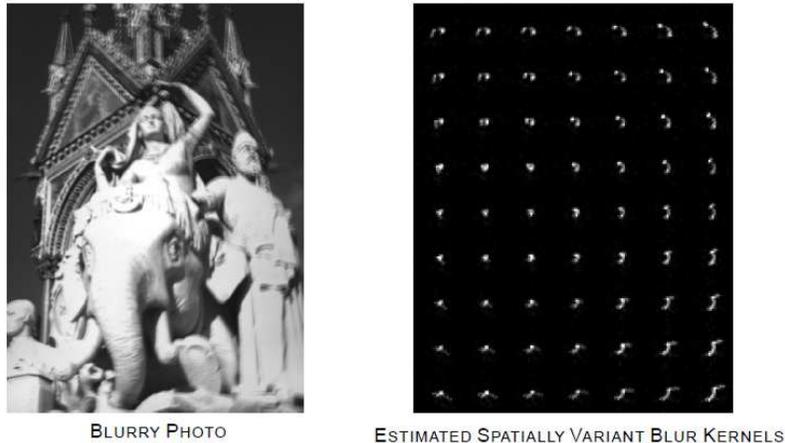}
\caption{{\bf Non-uniform blur example.} \emph{Left}: A blurry photo from \cite{HarmelingHS_NIPS10}. \emph{Right}: Estimated spatially-varying blur kernel array obtained using proposed method.  The resulting deblurred image is shown later in Figure~\ref{fig:Harmeling}.}
\label{fig:illus}
\end{figure}

While existing  algorithms for addressing non-uniform camera shake have displayed a measure of success, several important limitations remain.  First, some methods require either additional specialized hardware such as high-speed video capture~\cite{Tai09kasittr} or inertial measurement sensors~\cite{JoshiKZS10} for estimating motion, or else multiple images of the same scene~\cite{ChoCTL12}.  Secondly, even the algorithms that operate given only data from a single image typically rely on carefully engineered initializations, heuristics, and trade-off parameters for selecting salient image structure or edges, in part to avoid undesirable degenerate, no-blur solutions~\cite{Gupta10singleimage,HarmelingHS_NIPS10, HirschSHS_ICCV11, hu_bmvc2012}.  Consequently, enhancements and rigorous analysis may be problematic.
To address these shortcomings, we present an alternative blind deblurring algorithm built upon a simple, closed-form cost function that automatically discounts regions of the image that contain little information about the blur operator without introducing any additional salient structure selection steps.  This transparency leads to a nearly parameter free algorithm based upon a unique, adaptive sparsity penalty and provides theoretical arguments regarding how to robustly handle non-uniform degradations.  An example of estimated non-uniform or spatially-varying blur kernels is shown in Figure \ref{fig:illus}.

The rest of the paper is structured as follows.  Section~\ref{sec:related_works} briefly reviews relevant existing work on blind deblurring.
Section~\ref{sec:model_non_uni} then introduces the proposed non-uniform blind deblurring model, while theoretical justification and analyses are provided in Section~\ref{sec:Theoretical_analysis}.  Experimental comparisons with \emph{state-of-the-art} methods are carried out in Section~\ref{sec:Exp} followed by conclusions in Section~\ref{sec:con}.

\section{Related Work}
\label{sec:related_works}

Perhaps the most direct way of handling non-uniform blur is to simply partition the image into different regions and then learn a separate, uniform blur kernel for each region, possibly with an additional weighting function for smoothing the boundaries between two adjacent kernels.  The resulting algorithm admits an efficient implementation called \emph{efficient filter flow} \mbox{(EFF)}~\cite{Filter_Flow, HirschSSH10} and has been adopted extensively \cite{Leary, HarmelingHS_NIPS10, Xu_depth-awaremotion, non_uniform_restoration_chp3, JiHui12}.  The downside with this type of model is that geometric relationships between the blur kernels of different regions derived from the the physical motion path of the camera are ignored.

In contrast, to explicitly account for camera motion, the \emph{projective motion path} (PMP) model \cite{Tai09kasittr} treats a blurry image as the weighted summation of projectively transformed sharp images, leading to the revised observation model
\begin{eqnarray}\label{eq:obs_model_proj}
\begin{split}
\y = \sum_j w_j \PP_j\x +\n,
\end{split}
\end{eqnarray}
where $\PP_j$ is the $j$-th projection or homography operator (a combination of rotations and translations) and $w_j$ is the corresponding combination weight representing the proportion of time spent at that particular camera pose during exposure.  The uniform convolutional model can be obtained by restricting the general projection operators $\{\PP_j\}$ to be translations. In this regard, (\ref{eq:obs_model_proj}) represents a more general model that has been used in many recent non-uniform deblurring efforts~\cite{Tai09kasittr, Whyte_non-uniformdeblurring, Gupta10singleimage,hu_bmvc2012,ChoCTL12}.  PMP also retains the bilinear property of uniform convolution, meaning that
\begin{eqnarray}\label{eq:obs_model_proj_bilinear}
\begin{split}
\y &= \HH\x + \n = \D\w +\n,
\end{split}
\end{eqnarray}
where $\HH = \sum_j w_j \PP_j$ and $\D = [\PP_1\x, \PP_2\x, \cdots, \PP_j\x, \cdots]$ is a matrix of transformed sharp images.

The disadvantage of PMP is that it typically leads to inefficient algorithms because the evaluation of the matrix-vector product $\HH\x = \D\w$ requires generating many expensive intermediate transformed images.  However, EFF can be combined with the PMP model by introducing a set of basis images efficiently generated by transforming a grid of delta peak images~\cite{HirschSHS_ICCV11}.  The computational cost can be further reduced by using an active set for pruning out the projection operators with small responses~\cite{hu_bmvc2012}.  Furthermore, while the projective transformations (homographies) can generally involve six degrees-of-freedom (three for rotation and three for translation), recent work has demonstrated the effectiveness of using lower-dimensional restricted forms. For example, a 3D rotational camera motion model (i.e., using on roll, pitch, and yaw, no translations) is considered in~\cite{Whyte_non-uniformdeblurring}.  Likewise, a 3D camera motion model with $x,y$-translations and in-plane rotations has been used successfully in several other deblurring algorithms \cite{Gupta10singleimage,HirschSHS_ICCV11,hu_bmvc2012}. These two approximations display similar performance for sufficiently long focal lengths due to rotation-translation ambiguity in this setting~\cite{Gupta10singleimage}.

\section{A New Non-Uniform Deblurring Model} \label{sec:model_non_uni}

Following previous work~\cite{Fergus06removingcamera,LevinWDF11}, we work in the derivative domain of images for ease of modeling and better performance, meaning that $\x \in \mathbb{R}^m$ and $\y \in \mathbb{R}^n$ will denote the lexicographically ordered sharp and blurry image derivatives respectively.\footnote{The derivative filters used in this work are $\{[-1, 1], [-1, 1]^T\}$. Other choices are also possible.}  We will now derive a new non-uniform deblurring cost function followed by a majorization-minimization algorithm.

\subsection{Cost Function Derivation}

The observation model (\ref{eq:obs_model_proj}) is equivalent to the likelihood function
\begin{eqnarray}\label{eq:likelihood}
p(\y|\x, \w, \lambda) \propto \exp \left[ -\frac{1}{2\lambda} \Vert \y - \HH\x \Vert_2^2 \right].
\end{eqnarray}
Maximum likelihood estimation of $\x$ and $\w$ using (\ref{eq:likelihood}) is clearly ill-posed and so further regularization is required to constrain the solution space.  For this purpose, we adopt a sparse prior on $\x$ (in the gradient domain) as advocated in~\cite{Fergus06removingcamera,Whyte_non-uniformdeblurring}.  We assume the factorial form $p(\x)  = \prod_{i=1}^m p(x_i)$ for this prior, where
\begin{equation} \label{eq:convex_prior_xi}
p(x_i) = \max_{\gamma_i \geq 0} \ \mathcal{N}(x_i; 0,\gamma_i) \exp \left(-\frac{1}{2} f(\gamma_i) \right),
\end{equation}
which represents a weighted maximization over zero-mean Gaussians with different variances $\gamma_i$.  Here $f$ is some non-negative energy function, with different selections producing different priors on $\x$.  While it has been shown in \cite{Wipf_VEM_NIPS05} that any prior expressible in the form of (\ref{eq:convex_prior_xi}) will be super-Gaussian (sparsity promoting), we will rely on the special case where $f=0$ for our model.  This selection has been advocated in other applications of sparse estimation \cite{Wipf_Latent_Variable_TIT11}, has the advantage of being parameter free, and leads to a particularly compelling algorithm as will be shown below.

The hyperparameter variances $\gam = [\gamma_1,\ldots,\gamma_m]^T$ provide a convenient way of implementing several different estimation strategies \cite{Wipf_VEM_NIPS05}.  For example, perhaps the most straightforward is a form of MAP estimation given by
\begin{equation}
\max_{\x; \gam, \w\ge0} p(\y|\x, \w, \lambda) \prod_i \mathcal{N}(x_i; 0,\gamma_i),
\end{equation}
\noindent where simple update rules are available via coordinate ascent over $\x$, $\gam$, and $\w$ (a prior can also be included on $\w$ or $\lambda$ if desired).  However, recently it has been argued that an alternative estimation procedure may be preferred for canonical sparse linear inverse problems \cite{Wipf_Latent_Variable_TIT11}.  The basic idea, which naturally extends to the blind deconvolution problem, is to first integrate out $\x$, and then optimize over $\w$, $\gam$, as well as the noise level $\lambda$.  The final latent sharp image $\x$ can then be recovered using the estimated kernel and noise level along with standard non-blind deblurring algorithms. Later we will provide rigorous, independent rationalization for why the objective function produced through this process is ultimately superior to standard MAP.

Mathematically, this alternative estimation scheme requires that we solve
\begin{equation} \label{eq:type_II_cost_function}
\max_{\gam,\w,\lambda \geq 0} \int p(\y|\x,\w,\lambda) \mathcal{N}(x_i; 0,\gamma_i) d \x \equiv \min_{\gam,\w,\lambda \geq 0} \y^T \left(\HH \GM \HH^T + \lambda \II \right)^T \y + \log \left| \HH \GM \HH^T + \lambda \II \right|,
\end{equation}
where $\GM \triangleq \mbox{diag}[\gam]$.  While optimizing (\ref{eq:type_II_cost_function}) is possible using various general techniques such as the EM algorithm, it is computationally expensive in part because of the high-dimensional determinants involved with realistic-sized images.  Therefore, we instead minimize a convenient upper bound allowing us to circumvent this issue.  Specifically, using standard determinant identities we have
\begin{eqnarray} \label{eq:det_upper_bound}
\log \left| \HH \GM \HH^T + \lambda \II \right| & = &  n\log\lambda + \log|\GM| + \log \left| \lambda^{-1} \HH^T \HH + \GM^{-1} \right| \nonumber \\
& \leq &  n\log\lambda + \log|\GM| + \log \left| \lambda^{-1} \mbox{diag}\left[\HH^T \HH \right] + \GM^{-1} \right| \nonumber \\
& \equiv & \sum_i \log \left(\lambda + \gamma_i  \Vert \wbar_{i} \Vert^2_2  \right),
\end{eqnarray}
where $\HH = [ \wbar_1, \wbar_2, \ldots, \wbar_m]$.  Here $\wbar_i$ denotes the $i$-th column of $\HH$ and represents  the local blur kernel vector  associated with pixel location $i$ in the image plane. The local kernel $\wbar_i$ can be calculated by
\begin{eqnarray}
\wbar_i = \HH\ii_i = \sum_j w_j \PP_j \ii_i =  \B_i \w,
\end{eqnarray}
where $\ii_i$ denotes an all-zero image with a 1 at site $i$, and $\B_i \triangleq [\PP_1 \ii_i, \PP_2 \ii_i, \cdots, \PP_j \ii_i, \cdots]$.
Consequently we have $\Vert \wbar_i \Vert^2_2 = \w^T (\B_i^T \B_i)  \w$ for the norm embedded in (\ref{eq:det_upper_bound}).  The use of this diagonal approximation will not only make the proposed model computationally tractable, but it will also lead to an effective deblurring algorithm, as will be verified by the extensive experimental results in Section~\ref{sec:Exp}.

While optimizing (\ref{eq:type_II_cost_function}) using the bound from (\ref{eq:det_upper_bound}) can be justified in part using Bayesian-inspired arguments, the $\gam$-dependent cost function is far less intuitive than the standard penalized regression models dependent on $\x$  that are typically employed for blind deblurring.  However, using the framework from \cite{Wipf_Latent_Variable_TIT11}, it can be shown that the kernel estimate obtained by this process is formally equivalent to the one obtained via
\begin{eqnarray} \label{eq:cost_fun_non_uni}
 \begin{split}
 \min_{\x, \w\ge0, \lambda \geq 0} \frac{1}{\lambda} \Vert \y - \HH \x \Vert_2^2 +  \g(\x, \w, \lambda),
\end{split}
\end{eqnarray}
where
\begin{eqnarray} \label{eq:penalty}
 \begin{split}
\g(\x, \w, \lambda) \triangleq \sum_i g(x_i,\wbar_i,\lambda)
\end{split}
\end{eqnarray}
\noindent and
\begin{equation}\label{eq:g_full}
g(x_i, \wbar_i, \lambda ) \triangleq \frac{2|x_i| \|\wbar_i \|_2}{|x_i| \|\wbar_i \|_2 + \sqrt{4\lambda + x_i^2 \|\wbar_i \|_2^2}}  + \log \left(2\lambda +  x_i^2 \|\wbar_i \|_2^2 +  |x_i| \|\wbar_i \|_2 \sqrt{4\lambda + x_i^2 \|\wbar_i \|_2^2 }\right).
\end{equation}
The optimization from (\ref{eq:cost_fun_non_uni}) closely resembles a standard penalized regression (or equivalently MAP) problem used for blind deblurring.  The primary distinction is the penalty term $\g$, which jointly regularizes $\x$, $\HH$, and $\lambda$.  We discuss a tractable algorithm for optimization that accounts for this intrinsic coupling in Section \ref{sec:algorithm}, followed by analysis in Section \ref{sec:Theoretical_analysis}.

\begin{algorithm}[t]
\caption{Non-Uniform Blind Deblurring.}
\begin{algorithmic}[1]
\STATE {\bf Input:} {a blurry image $\y$}
\STATE {\bf Initialize:}   blur parameter vector ${\w}$, noise level $\lambda$, and $d = n 10^{-4}$
\STATE  {\bf While} stopping criteria is not satisfied, do \vspace*{0.4cm}
{
        \begin{itemize}
        \item    {\bf Image Update:} \\
         $ \x \leftarrow  [\frac{{\HH}^T {\HH}}{{\lambda}} +  {\GM}^{-1} ]^{-1}  \frac{{\HH}^T \y}{{\lambda}}$\quad  where $\GM \triangleq \mbox{diag}[\gam]$ \vspace*{0.7cm}

        \item
            {\bf Latent Representation Update:}\\
               $
                {\gamma_i} \leftarrow {x_i}^2 + z_{i}$,  \mbox{ with }  $z_{i} =  \frac{1}{\frac{\Vert \wbar_i  \Vert^2_2}{\lambda} + \gamma_i^{-1}}$   \vspace*{0.4cm}

     \item {\bf Blur Update:} \\
       $ \w \leftarrow \arg \min_{\w\ge 0} \ \Vert \y - \D \w\Vert_2^2 +   \w^T \left(\sum_i z_i \B_i^T\B_i \right) \w $  \vspace*{0.4cm}

 \item {\bf Noise Level Update:} \\
$ {\lambda} = \frac{\Vert \y - \HH \x \Vert_2^2 + \beta +d}{n}$, \mbox{ with } $\beta = \sum_i \frac{\Vert \wbar_i \Vert^2_2}{\frac{\Vert \wbar_i \Vert^2_2}{\lambda} + \gamma_i^{-1}}$   \vspace*{0.4cm}

        \end{itemize}

}
\STATE {\bf End}
\end{algorithmic}
\label{algo:algo_non_uniform_BDB}
\end{algorithm}

\subsection{Minimization Algorithm}  \label{sec:algorithm}

The proposed practical blind deblurring algorithm simply involves solving (\ref{eq:cost_fun_non_uni}).  This can be accomplished by instead minimizing a convenient upper bound $\mathcal{L}(\x, \w, \gam, \lambda)$ defined as
\begin{eqnarray}\label{eq:cost_fun_upperbound}
\mathcal{L}(\x, \w, \gam, \lambda) \triangleq \frac{1}{\lambda} \Vert \y - \HH \x \Vert_2^2 +  \sum_i \left[\frac{x_i^2}{\gamma_i}  + \log(\lambda + \gamma_i \Vert  \wbar_i \Vert^2_2)\right],
\end{eqnarray}
where $\gam \triangleq [\gamma_1,\ldots,\gamma_m]^T$ is a vector of latent variables controlling the shape of the bound.  The form of (\ref{eq:cost_fun_upperbound}) is motivated by the fact that the proposed penalty function satisfies
\begin{equation} \label{eq:g_in_variational_form}
g(x_i,\w,\lambda) = \min_{\gamma_i \geq 0}  \frac{x_i^2}{\gamma_i}  + \log(\lambda + \gamma_i \Vert  \wbar_i \Vert^2_2).
\end{equation}
This expression can be shown by optimizing over $\gamma_i$, plugging in the resulting value which can be obtained in closed-form, and then simplifying. From (\ref{eq:g_in_variational_form}) it then follows that
\begin{equation}
\mathcal{L}(\x, \w, \gam, \lambda) \geq \frac{1}{\lambda} \Vert \y - \HH \x \Vert_2^2 +  \g(\x, \w, \lambda)
\end{equation}
for all $\gam \geq 0$, with equality when each $\gamma_i$ solves (\ref{eq:g_in_variational_form}).  Consequently we can solve (\ref{eq:cost_fun_non_uni}) iteratively by minimizing $\mathcal{L}(\x, \w, \gam, \lambda)$ in an alternating fashion over $\x$, $\w$, $\gam$, and $\lambda$.  This majorization-minimization technique~\cite{CCCP_NIPS03,Wipf_Latent_Variable_TIT11} has similar convergence properties to the EM algorithm.  The resulting procedure is summarized in Algorithm~\ref{algo:algo_non_uniform_BDB}.  The details of each constituent subproblem are derived in Appendix~\ref{apd:algo_derivation}.

Algorithm~\ref{algo:algo_non_uniform_BDB} is very straightforward. The image and blur are updated by solving simple quadratic minimization problems. The update rules for the latent variables $\gam$ and the noise level $\lambda$ are also minimally complex.

For simplicity in practice,  we have only used projection operators $\PP_j$ involving in-plane translations and rotations similar to~\cite{Gupta10singleimage} for modeling the camera shake, and use the EFF model \cite{HarmelingHS_NIPS10} for reducing the computational expense.
We have also incorporated the technique similar to the one used in \cite{hu_bmvc2012}, whereby irrelevant projection operators are pruned out while some new ones are added by sampling around the remaining projections using a Gaussian distribution with a small variance.   Note that this heuristic is only for reducing the computational complexity; using the fully sampled basis set would generate the best results.  Also, a standard multi-scale estimation scheme is incorporated consistent with most recent blind deblurring work \cite{Fergus06removingcamera,LevinWDF11, Whyte_non-uniformdeblurring, HirschSHS_ICCV11}.

Finally, we emphasize that Algorithm \ref{algo:algo_non_uniform_BDB} only provides an estimate of $\x$ in the gradient domain.  Consequently, consistent with other methods we use the estimated blur parameters $\w$ in a final non-blind deconvolution step to recover the latent sharp image.

\section{Theoretical Analysis}\label{sec:Theoretical_analysis}

The proposed blind deblurring strategy involves simply minimizing (\ref{eq:cost_fun_non_uni}); no additional steps for structure or salient edge detection are required unlike other state-of-the-art approaches.  This section will examine theoretical properties of the proposed penalty function $\g$ embedded in (\ref{eq:cost_fun_non_uni}) that ultimately allows such a simple algorithm to succeed.  We note that, unlike  prototypical penalized sparse regression models used for blind deblurring, $\g$ is non-separable with respect to the image $\x$ and the blur parameters $\w$, meaning that it cannot be decomposed as $\g(\x, \w, \lambda) = h_1(\x) + h_2(\w)$ for some functions $h_1$ and $h_2$.  Moreover, it also depends on the noise level $\lambda$, a novel dependency with important consequences as shown below.

To address these distinctions, we will examine $\g$ from two complementary perspectives.  First, we will treat $\g$ as a function of $\x$ parameterized by $\w$ and $\lambda$, and then subsequently we will treat it as a function of $\w$ parameterized by $\x$ and $\lambda$.  This will ultimately serve to demonstrate that the intrinsic coupling is highly advantageous over any separable functions $h_1$ and $h_2$.

\subsection{The Effective Penalty on $\x$}

For analysis purposes we first introduce the definition of \emph{relative concavity}~\cite{RC_Palmer}.
\begin{Def}[Relative Concavity]\label{def:1}
Let $\gfun$ be a strictly increasing function on $[a, b]$.  The function
$\ffun$ is \textbf{concave relative} to $\gfun$ on the interval $[a,b]$ if and only if  $\ffun(y) \le \ffun(x) + \frac{\ffun'(x)}{\gfun'(x)} \left[ \gfun(y)-\gfun(x) \right]$
holds $\forall x,y \in [a, b]$.
\end{Def}

We will use $\ffun \prec \gfun$ to denote that $\ffun$ is  concave relative to $\gfun$ on $[0,\infty)$.  This can be understood as a natural generalization of the traditional notion of a concavity, in that a concave function is equivalently \emph{concave relative to a linear function} per Definition \ref{def:1}.  In general, if $\ffun \prec \gfun$, then when $\ffun$ and $\gfun$  are set to have the same  functional value and the same slope at any given point (i.e., by an affine transformation of $\gfun$), then $\ffun$ lies completely under $\gfun$.

\begin{figure}
\centering
\includegraphics[height=6cm]{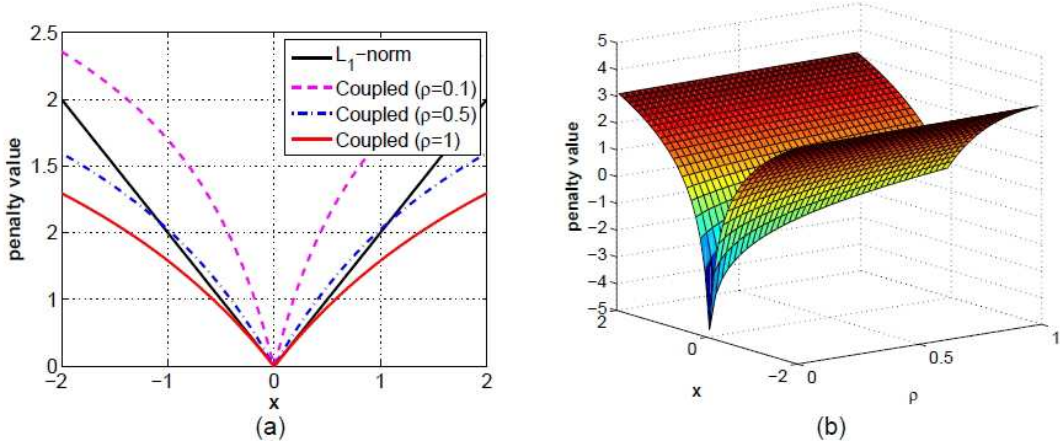}
\caption{(a) A 1D example of the coupled penalty $h(|x|, \rho)$ (normalized) with different $\rho$ values. The $\mathcal{\ell}_1$ norm is included for comparison. (b) A 2D example surface plot of the coupled penalty function $h(|x|, \rho)$.}
\label{fig:penalty_fun}
\end{figure}

Now consider the function $h(\cdot; \rho) :\mathbb{R}^+ \rightarrow \mathbb{R}$ defined as
\begin{equation}\label{eq:x_only_penalty}
h(z; \rho ) \triangleq \frac{2z }{z  + \sqrt{4\rho + z^2 }}  + \log \left(2\rho +  z^2  +  z \sqrt{4\rho + z^2 }\right).
\end{equation}
It follows then that
\begin{equation} \label{eq:reexpress_g}
\g(\x,\w,\lambda) = \sum_i h(|x_i|; \rho_i) + \sum_i 2\log \|\wbar_i \|_2
\end{equation}
where $\rho_i \triangleq  \lambda/\Vert \wbar_i  \Vert^2_2$.  The second summation in (\ref{eq:reexpress_g}) is independent of $\x$, so here we will focus on the first term, which suggests that the  penalty function shape over the image $\x$ depends only on this ratio of noise level to the squared norm of the local kernel.  This leads to some desirable properties relevant to blind deblurring.  Figure~\ref{fig:penalty_fun} shows how $h(|x|;\rho)$ varies its shape with $\rho$.

\vspace*{0.4cm}

\begin{Thm}[Spatially-Adaptive Sparsity] \label{thm:spa_adap_sparsity}

The proposed penalty $h$ satisfies:
\begin{enumerate}
\item $h(z; \rho)$ is a concave, non-decreasing function of $z$ for all $\rho \ge 0$.
\item If $\rho_1<\rho_2$, then $\frac{\partial h(z; \rho_1)}{\partial z} > \frac{\partial h(z; \rho_2)}{\partial z}$ and $h(z; \rho_1) \prec h(z; \rho_2)$.
\end{enumerate}
\end{Thm}

\vspace*{0.4cm}

The proof has been deferred to Appendix~\ref{apd:proof}.  The first property of Theorem~\ref{thm:spa_adap_sparsity} implies that the derived penalty function $h$ favors solutions $\x$ with some entries exactly equal to zero.\footnote{When $\w$ and $\lambda$ are fixed, optimization of $\x$ using (\ref{eq:cost_fun_non_uni}) falls into the canonical form $\min_{\x} \|\y - \A \x \|_2^2 + \lambda \sum_i \psi(|x_i|)$, where $\psi$ is a concave, non-decreasing function by virtue of Theorem \ref{thm:spa_adap_sparsity}.  Such problems will provably have some elements of $\x$ equal to zero if $\A$ is overcomplete and/or if $\lambda$ is sufficiently large \cite{RaoECPK03,Wipf_Latent_Variable_TIT11}.}  In this sense it is similar to more traditional penalty functions based on the $\ell_p$ pseudo-norm $\sum_i |x_i|^p$, $p \in (0,1]$, or other related sparsity measures.  Consequently, the more unique attributes of $h$ stem from the second property of Theorem~\ref{thm:spa_adap_sparsity}, which leads to a desirable spatially-adaptive form of sparsity.

To understand the significance of these properties, it helps to review an important practical consideration involved when designing robust deblurring systems.  First, blind deconvolution algorithms applied to deblurring are heavily dependent on some form of stagewise coarse-to-fine approach, whereby the blur operator is repeatedly re-estimated at successively higher resolutions.  At each stage, a lower resolution version is used to initialize the estimate at the next higher resolution.  One way to implement this approach is to initially use large values of $\lambda$ such that only dominant, primarily low-frequency image structures dictate the optimization \cite{LevinWDF11_PAMI}.  During subsequent iterations as the blur operator begins to reflect the correct coarse shape, $\lambda$ can be gradually reduced to allow the recovery of more detailed, fine structures.

A highly sparse (concave) prior can ultimately be more effective in differentiating sharp images and fine structures than a convex one. Detailed supported evidence for this claim can be found in \cite{Fergus06removingcamera, sps_deblur, KrishnanF09_NIPS, Cho_PAMI}. However, if such a prior is applied at the initial stages of estimation, the iterations are likely to become trapped at suboptimal local minima, of which there will always be a combinatorial number.  Moreover, in the early stages, the effective noise level is actually high due to errors contained in the estimated blur kernel, and exceedingly sparse image penalties are likely to produce unstable solutions.

Theorem \ref{thm:spa_adap_sparsity} implies that the proposed method may implicitly avoid these problems by initializing with a large $\lambda$ (and therefore a large $\rho$), such that the penalty function is initially nearly convex in $|x_i|$ at all pixels $i$.  As the iterations proceed and coarse structures are resolved, the effective noise level (or modeling error) reduces, along with the estimated $\lambda$.  Consequently, later when fine structures need to be resolved, the penalty function becomes less convex as $\lambda$ is automatically reduced by the learning process, but the risk of local minima and instability is ameliorated by the fact that we are likely to be already in the neighborhood of a desirable basin of attraction.

The form of image penalty adaptation just described occurs globally across all pixels.  However, a more interesting and nuanced shape adaptation occurs regionally based on differences in the local blur estimate $\wbar_i$, which also affects the pixel-wise parameter $\rho_i$.  Recall that $\wbar_i$ can be viewed as the local blur kernel around pixel $i$, meaning that in this local region the blurry image can be roughly modeled as $\wbar_i \ast \x$, where $\ast$ denotes the standard 2D convolution.  Given the feasible simplex $\w \geq 0$ and $\sum_i w_i = 1$ commonly assumed for blind deblurring, it can be shown that $1/L \leq \| \wbar_i \|_2^2 \leq 1$, where $L$ is the maximum number of pixels in any local kernel.  The upper bound is achieved when the local kernel is a delta solution, meaning only one nonzero element and therefore minimal blur.  This scenario produces the highest relative concavity (i.e., sparsity) by virtue of  Theorem \ref{thm:spa_adap_sparsity} since $\rho_i$ will be minimized.  Such a high degree of sparsity is warranted here because there is little risk of local minima in regions with such a simple kernel and the added concavity can help to differentiate small-scale structures necessary for obtaining a globally reasonable solution.  Note also that if we estimate the correct $\w$ based on a few local regions, then the overall deblurred image will be sharp to the extent that our forward model is correct.

In contrast, the lower bound on $\| \wbar_i \|_2^2$ occurs when every element of $\wbar_i$ has an equal value.  Now $\rho_i$ is maximized and the relative concavity is minimal, meaning $h(|x_i|; \rho_i)$ is the closest to being convex.  Again, this represents a desirable tuning mechanism.  A uniformly distributed $\wbar_i$ indicates maximal blur, and therefore higher risk for local minima.  Moreover, in such regions, only dominate edges/structures will remain, and so a nearly convex penalty is sufficient for disambiguation of residual coarse details.  Moreover, because of property two, not only is $h$ nearly convex, but its slope is also minimal, meaning the influence to the overall cost function is also minimized.  Consequently, regions with smaller local blur kernels and significant edges will automatically dominate the image penalty, while flat regions or areas with large blur will be discounted.  Importantly, this spatially-adaptive sparsity occurs without the need for additional structure selection measures, meaning carefully engineered heuristics designed to locate prominent edges such that good global solutions can be found with minimally non-convex image penalties \cite{hqdeblurring_siggraph2008,fast_motion_deblur_2009, XuJ10_ECCV,HarmelingHS_NIPS10,HirschSHS_ICCV11}.

\begin{figure*}[t]
\centering
\includegraphics[width=16cm]{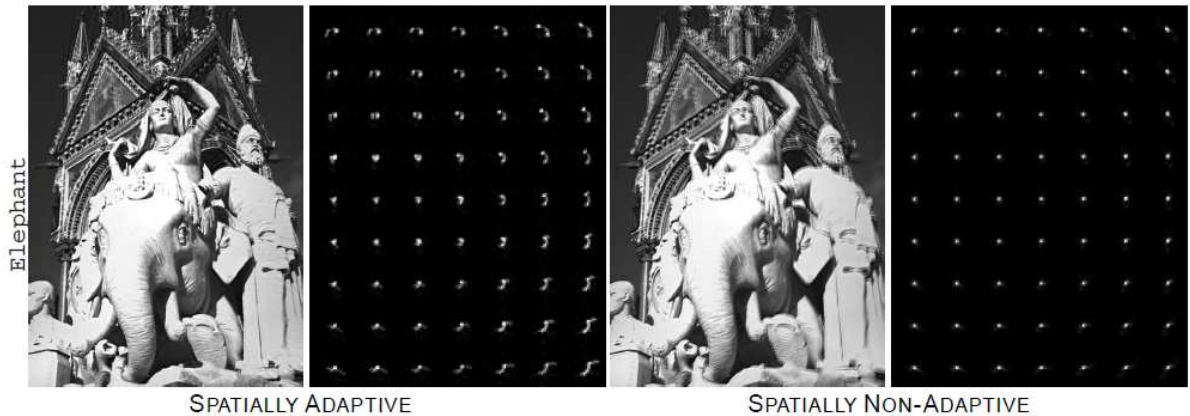}
\caption{{\bf Effectiveness of spatially adaptive sparsity.} The deblurred images and estimated kernel maps obtained using the proposed spatially adaptive sparsity (\emph{left}) and standard spatially non-adaptive sparsity (\emph{right}) for the {\ttfamily Elephant} image shown in Figure~\ref{fig:Harmeling}.}
\label{fig:effects_non_uni}
\end{figure*}

\begin{figure}[t]
\centering
\includegraphics[width=16cm]{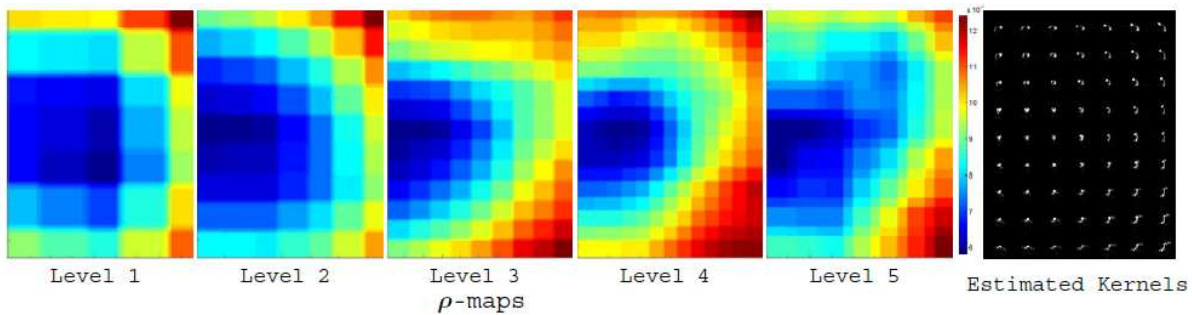}
\caption{Map of $\brho$ as estimated by the proposed algorithm at different resolutions using the blurry {\ttfamily Elephant} image shown in Figure~\ref{fig:Harmeling}.  This map reflects the degree of local blurring which ultimately controls the shape of the image penalty.}
\label{fig:kernel_norm_map_bar}
\end{figure}

Figure~\ref{fig:effects_non_uni} presents example deblurring results on the real-world {\small\ttfamily Elephant} image from \cite{HarmelingHS_NIPS10} both with and without the described spatially-adaptive sparsity mechanism.  We also display the corresponding image of $\brho \triangleq [\rho_1,\ldots, \rho_m]$ values in Figure~\ref{fig:kernel_norm_map_bar}, which determines which regions of the estimated sharp image will have the greatest impact on the cost function for the spatially-adaptive case. For purely uniform blur, this $\brho$-map would be constant neglecting small boundary effects, while for rotational blur, it would be smallest at the rotation center, and larger on the periphery.  The learned $\brho$-map from a real image is refined across the coarse-to-fine hierarchy and reflects a combination of rotations and translations, modulating the relatively concavity using the inverse of the estimated local kernel spread.  Importantly, if we remove this spatial adaptation, and instead substitute the fixed norm $\| \w \|_2^2$ for all $i$, the performance degrades as shown in Figure~\ref{fig:effects_non_uni}.

\subsection{The Effective Penalty on $\w$}

We may also consider $\g$ as a function of $\w$ with shape modulated by $\x$ and $\lambda$ as well as the basis functions $\B_i$, leading to an interesting, complementary perspective.  With this intent in mind, we define
\begin{equation}\label{eq:w_only_penalty}
\nu(\w; \mu, \B ) \triangleq  \frac{2 \mu \|\w\|_{\B}}{\mu \|\w\|_{\B} + \sqrt{4 + \mu^2 \|\w\|_{\B}^2}}  +  \log \left(2 +  \mu^2 \|\w\|_{\B}^2 +  \mu \|\w\|_{\B} \sqrt{4 + \mu^2 \|\w\|_{\B}^2 }\right).
\end{equation}
where $\| \w\|_{\B}$ denotes the weighted quadratic norm $\sqrt{\w^T (\B^T \B) \w}$ (and so it follows that $\| \w\|_{\B_i} = \| \wbar_i \|_2$).  By definition we then have
\begin{equation}
\g(\x,\w,\lambda) = \sum_i \nu(\w; \mu_i, \B_i) + m \log \lambda,
\end{equation}
where $\mu_i \triangleq  |x_i|/\sqrt{\lambda}$.  Note that because many $x_i$ may equal zero (in regions with zero gradient), we must define the shape parameters $\mu_i$ differently from the previous section. Moreover, while certain symmetries exist between the effective penalties on $\x$ and $\w$, the analysis and interpretations turn out to be significantly divergent.

Interestingly, $\nu$ is completely blind to image regions determined to be relatively flat.  More specifically, if a gradient $x_i$ is zero, then $\mu_i$ is zero and $\nu(\w; 0, \B_i) = 0$ contributes no penalty on $\w$.  Consequently, as estimation proceeds and coarse image gradient estimates become available, the blur operator penalty is increasingly dominated by edges and structured image areas.  However, it is important to examine how the shape of $\nu$ changes depending on where and how these edges are distributed relative to local blurring as dictated by each $\B_i$.

Simply put, if the majority of large gradients occur near the center of some rotations, then the penalty will provably become nearly flat.  This occurs because the corresponding $\B_i$ for such regions will be nearly a zero matrix with a single row of ones (and if location $i$ is directly in the rotation center, it will be exactly so).  Given the constraint $\sum_i w_i = 1$, any feasible $\w$ will then necessarily produce almost the same $\| \w \|_{\B_i}$ value, and hence $\nu$ will be relatively flat.  This is consistent with the intuition that minimal kernel regularization is required when there is limited blurring of the primary edges.

In contrast, when the majority of significant gradients occur in areas with large local blurring (as a combination of translations and distant rotations), then the kernel penalty will impose strong quadratic regularization on $\w$.  This can be explained by noting that $\B_i^T \B_i$ will be approximately an identity matrix for translations (ignoring boundary effects) and distant rotations (which behave like translations far from the rotation center).  This is also desirable consequence since a relatively diffuse blurring operator will be needed to resolve such edges.

Thus ultimately, the penalty on $\w$ transitions between a form of quadratic regularizer, which favors many nonzero elements of $\w$ suitable for characterizing larger blur, and no penalty at all (within the specified constraint set).  Moreover, this adaptive regularization is processed using a non-linearity in $\nu$ such that data-fit and kernel penalties are properly balanced.  By this we mean that if the image gradients $\x$ are scaled by some factor $\alpha$ (i.e., $\x \rightarrow \alpha \x$), then the $\w$ which solves
\begin{equation} \label{eq:w_only_cost}
\min_{\w} \Vert \y - \D \w\Vert_2^2 + \sum_i \nu(\w; \mu_i,\B_i),
\end{equation}
will simply be scaled by the same factor $\alpha$. Because (\ref{eq:w_only_cost}) represents the cost function from (\ref{eq:cost_fun_non_uni}) with $\x$ fixed, this form of invariance helps to explain why the proposed algorithm is largely devoid of trade-off parameters that are typically used to calibrate the kernel penalty.  Note that \emph{both} the nonlinearity with respect to the norms $\| \w \|_{\B_i}$ \emph{and} the $x_i$-dependency in $\nu$ contribute to this invariance while simultaneously maintaining an integrated cost function over both $\x$ and $\w$ (which is easily shown to be globally scale invariant as well).

\section{Experiments}
\label{sec:Exp}

This section compares the proposed method with several \emph{state-of-the-art} algorithms for both uniform and non-uniform blind deblurring using  real-world images.

\subsection{Uniform Deblurring}
Any non-uniform deblurring approach should naturally reduce to an effective uniform algorithm when the blur transformations are simply in-plane translations.  We first evaluate our algorithm in the uniform case where existing benchmarks facilitate quantitative comparisons with state-of-the-art methods.  For this purpose we reproduce the experiments from \cite{LevinWDF11} using the benchmark test data from~\cite{LevinWDF11_PAMI},\footnote{{\url{http://www.wisdom.weizmann.ac.il/~levina/papers/LevinEtalCVPR09Data.rar}}}  which consists of 4 base images of size $255\times 255$ and 8 different blurring effects, leading to a total of 32 blurry images.
Ground truth blur kernels were estimated by recording the trace of focal reference points on the boundaries of the sharp images.
The kernel sizes range from $13\times 13$ to $27\times 27$.
We compare the proposed method with only in-plane translation,  with the algorithms of Shan \emph{et al.}~\cite{hqdeblurring_siggraph2008}, Xu \emph{et al.}~\cite{XuJ10_ECCV}, Cho \emph{et al.}~\cite{fast_motion_deblur_2009}, Fergus \emph{et al.}~\cite{Fergus06removingcamera} and Levin \emph{et al.}~\cite{LevinWDF11}.

\begin{figure}[t]
\centering
\includegraphics[height=6cm]{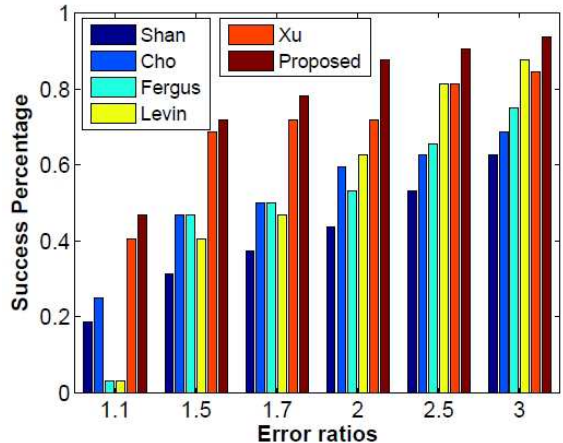}
\caption{Evaluation of uniform deblurring results using cumulative histogram of the deconvolution error ratios across 32 test examples from~\cite{LevinWDF11_PAMI}. The height of the bar indicates the percentage of images having error ratio below that level. Higher bars indicate better performance.}
\label{fig:acc_hist}
\end{figure}

The SSD (Sum of Squared Difference) metric defined in~\cite{LevinWDF11_PAMI} is used for measuring the error between estimated and the ground-truth images.
To normalize for the fact that a harder kernel gives a larger image reconstruction error even when the true kernel is known (because the corresponding non-blind deconvolution problem is also harder), the SSD ratio between the image deconvolved with the estimated kernel and the image deconvolved with the ground-truth kernel is used as the final evaluation measure. The cumulative histogram of the error ratios is shown in Figure~\ref{fig:acc_hist}.
The height of the bar indicates the percentage of images having error ratio below that level. Higher bars indicate better performance, revealing that the proposed method significantly outperforms existing methods on uniform deblurring tasks.

\begin{figure*}[t]
\centering
\includegraphics[width=16cm]{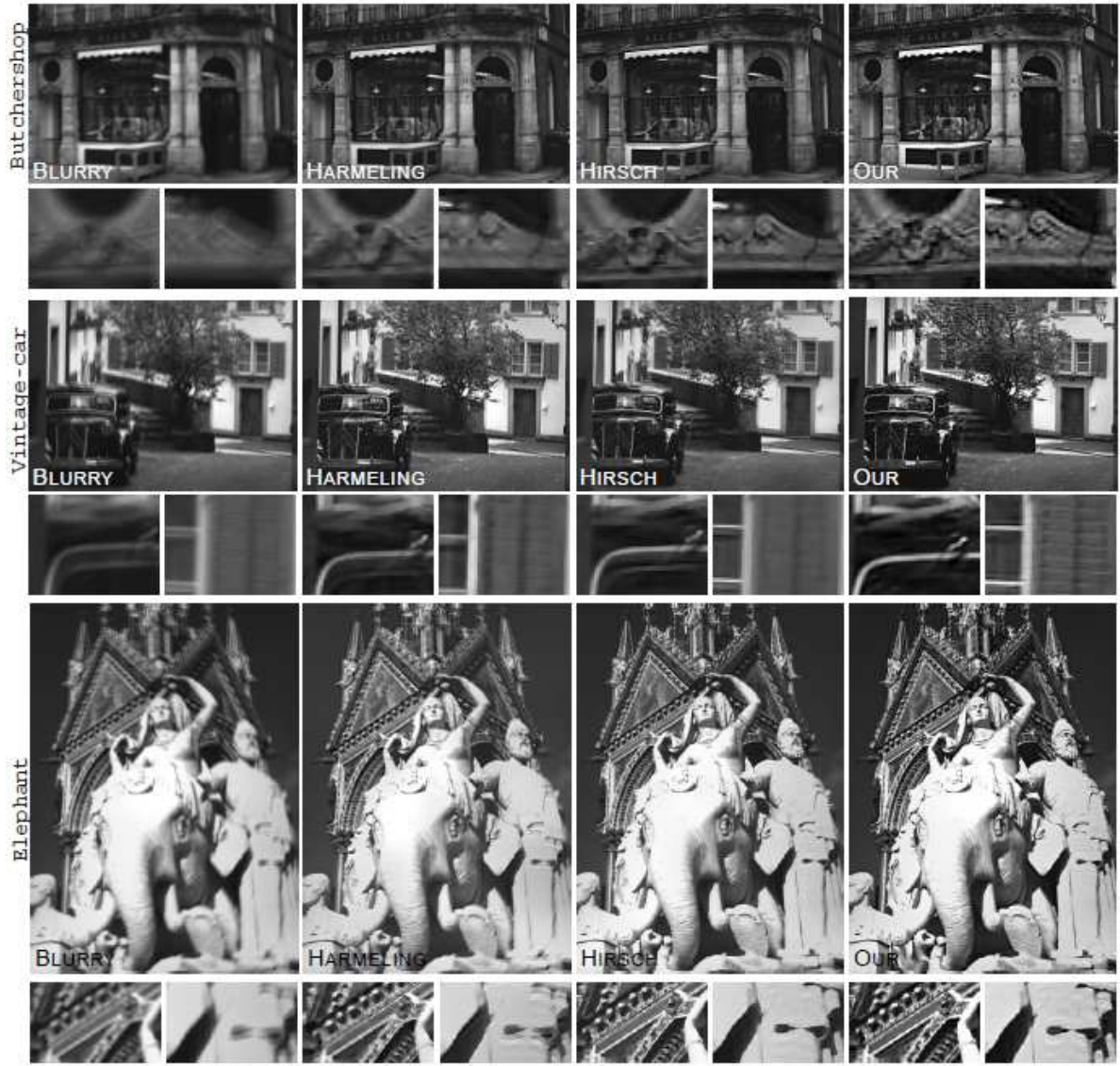}
\caption{Non-uniform deblurring comparisons with Harmeling \emph{et al.}~\cite{HarmelingHS_NIPS10} and Hirsch \emph{et al.}~\cite{HirschSHS_ICCV11} using the three real-world test images {\ttfamily Butchershop}, {\ttfamily Vintage-car}, and {\ttfamily Elephant} provided in~\cite{HarmelingHS_NIPS10}.  Additionally, ground-truth local blur kernels associated with each of these deblurring results are shown in Figure~\ref{fig:kernel_pattern} below.}
\label{fig:Harmeling}
\end{figure*}

\begin{figure*}[t]
\centering
\includegraphics[width=16cm]{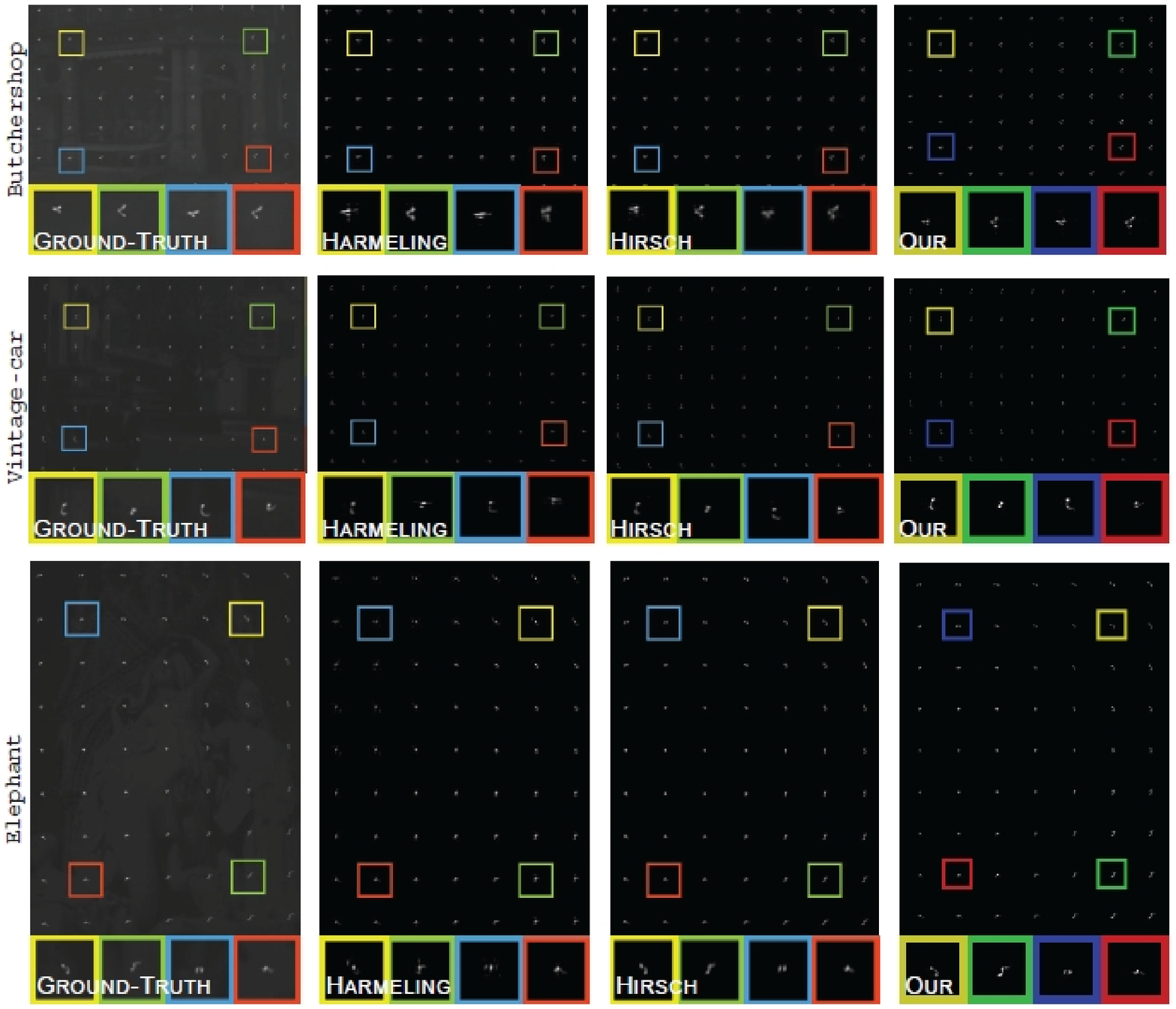}
\caption{Non-uniform kernel estimation comparisons associated with the {\ttfamily Butchershop}, {\ttfamily Vintage-car}, and {\ttfamily Elephant} images from Figure \ref{fig:Harmeling} above.  Note that kernels have been resized for display purposes.}
\label{fig:kernel_pattern}
\end{figure*}

\subsection{Non-uniform Deblurring}

For non-uniform deblurring, quantitative comparisons are much more difficult because of limited benchmark data with available ground truth.  Moreover, because source code for most state-of-the-art non-uniform algorithms is not available, it is not feasible to even qualitatively compare all methods across a wide range of images.  Consequently, the only feasible alternative is simply to visually compare our algorithm using images contained in previously published papers with the deblurring results presented in those papers.  In this context, a successful non-uniform blind deblurring algorithm is one that consistently performs comparably or better than all existing algorithms on the respective images where these algorithms have been previously tested.  This section strongly suggests that the proposed approach is such a successful algorithm, even without any effort to optimize the non-blind deconvolution step (which is required after kernel estimation as mentioned previously).

{\bf Comparisons with  Harmeling \emph{et al.}~\cite{HarmelingHS_NIPS10} and Hirsch \emph{et al.}~\cite{HirschSHS_ICCV11}:} Figure~\ref{fig:Harmeling} displays deblurring comparisons based on the {\small\ttfamily Butchershop}, {\small\ttfamily Vintage-car}, and {\small\ttfamily Elephant} images provided in~\cite{HarmelingHS_NIPS10}.  Overall, the proposed algorithm typically reveals more fine details than the other methods, despite its simplicity and lack of salient structure selection heuristics or trade-off parameters.\footnote{Results throughout this section are better viewed electronically with zooming.}  Note that with these three images, ground truth blur kernels were independently estimated using a special capturing process.  See \cite{HarmelingHS_NIPS10} for more details on this process.  As shown in the Figure~\ref{fig:kernel_pattern} using a $7 \times 9$ (or $9 \times 7$) array for visualization, the estimated blur kernel patterns obtained from our algorithm are generally better matched to the ground truth relative to the other methods, a performance result that compensates for any differences in the non-blind step.

\begin{figure*}[t]
\centering
\includegraphics[width=16cm]{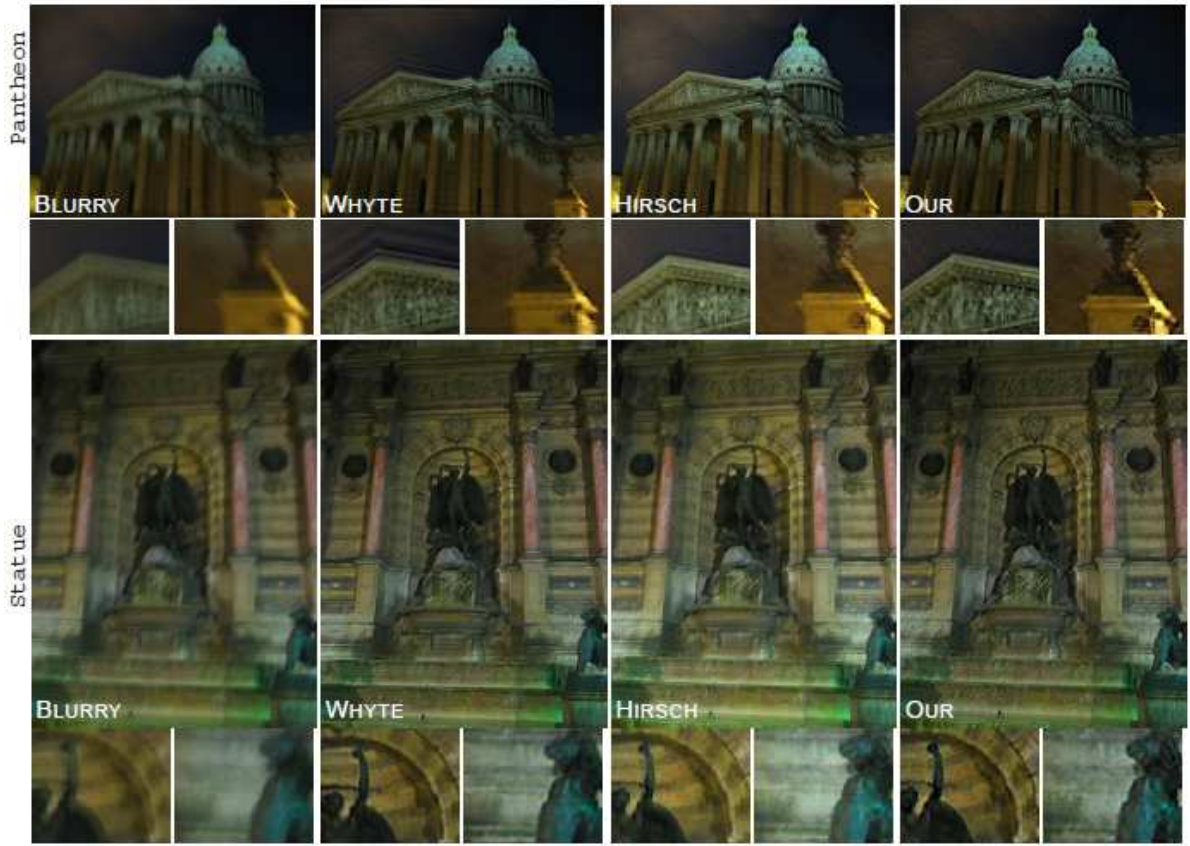}
\caption{Non-uniform deblurring comparisons with  Whyte \emph{et al.}~\cite{Whyte_non-uniformdeblurring} and Hirsch \emph{et al.}~\cite{HirschSHS_ICCV11} on the real-world images {\ttfamily Pantheon} and {\ttfamily Statue} from~\cite{Whyte_non-uniformdeblurring}.}
\label{fig:Whyte}
\end{figure*}

{\bf Comparisons with Whyte \emph{et al.}~\cite{Whyte_non-uniformdeblurring} and Hirsch \emph{et al.}~\cite{HirschSHS_ICCV11}:} We further evaluate our algorithm using the {\small\ttfamily Pantheon} and {\small\ttfamily Statue} images from~\cite{Whyte_non-uniformdeblurring}.  Results are shown in Figure~\ref{fig:Whyte}, where we observe that the deblurred image from Whyte \emph{et al.}~has noticeable ringing artifacts.  In contrast, our result is considerably cleaner.  On the {\small\ttfamily Pantheon} example the deblurring result from Whyte \emph{et al.} has significant ringing artifacts while the result from Hirsch \emph{et al.} seems to be suffering from some chrome distortions as indicated by the dome area of the pantheon. Our result on the other hand, has very few artifacts or chrome distortions.  On the {\small\ttfamily Statue} image the result of Whyte \emph{et al.} is generated using a blurry image paired with another additional noisy image of the same scene captured with a shorter exposure time length.  Our method and that of Hirsch \emph{et al.}, without the benefit of such additional image data, can nonetheless generate a deblurring result with comparable quality.

\begin{figure*}[t]
\centering
\includegraphics[width=16cm]{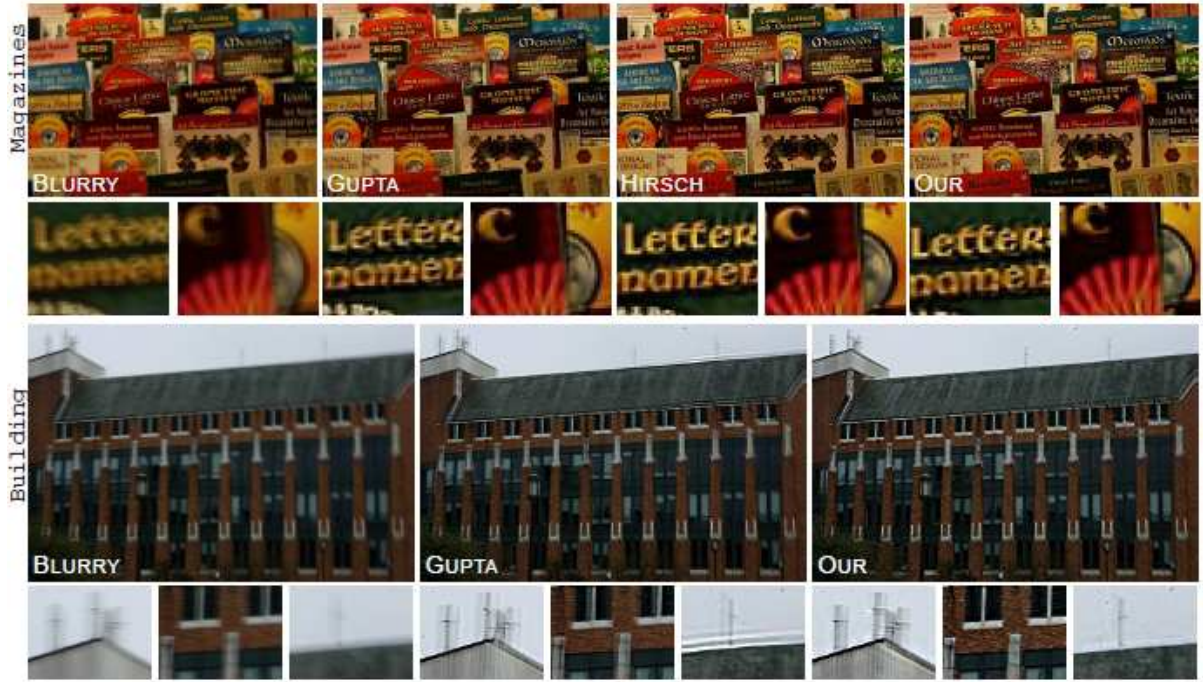}
\caption{Non-uniform deblurring comparisons with Gupta \emph{et al.}~~\cite{Gupta10singleimage} and Hirsch \emph{et al.}~~\cite{HirschSHS_ICCV11} on the real-world images   {\ttfamily Magazines} and {\ttfamily Building} from~\cite{Gupta10singleimage}.  Note that Hirsch \emph{et al.}~do not provide a deblurring result for the {\ttfamily Building} image.}
\label{fig:Gupta}
\end{figure*}

{\bf Comparisons with Gupta \emph{et al.}~\cite{Gupta10singleimage} and Hirsch \emph{et al.}~\cite{HirschSHS_ICCV11}:} We next experiment using the test images {\small\ttfamily Magazines} and {\small\ttfamily Building} from~\cite{Gupta10singleimage}, which contain large, challenging rotational blur effects.  Figure~\ref{fig:Gupta} reveals that our algorithm contains fewer artifacts and more fine details relative to Gupta \emph{et al.}, and comparable results to Hirsch \emph{et al.} on the {\small\ttfamily Magazines} image.  Note that Hirsch \emph{et al.} do not provide a deblurring result for the {\small\ttfamily Building} image.

\begin{figure*}[t]
\centering
\includegraphics[width=16cm]{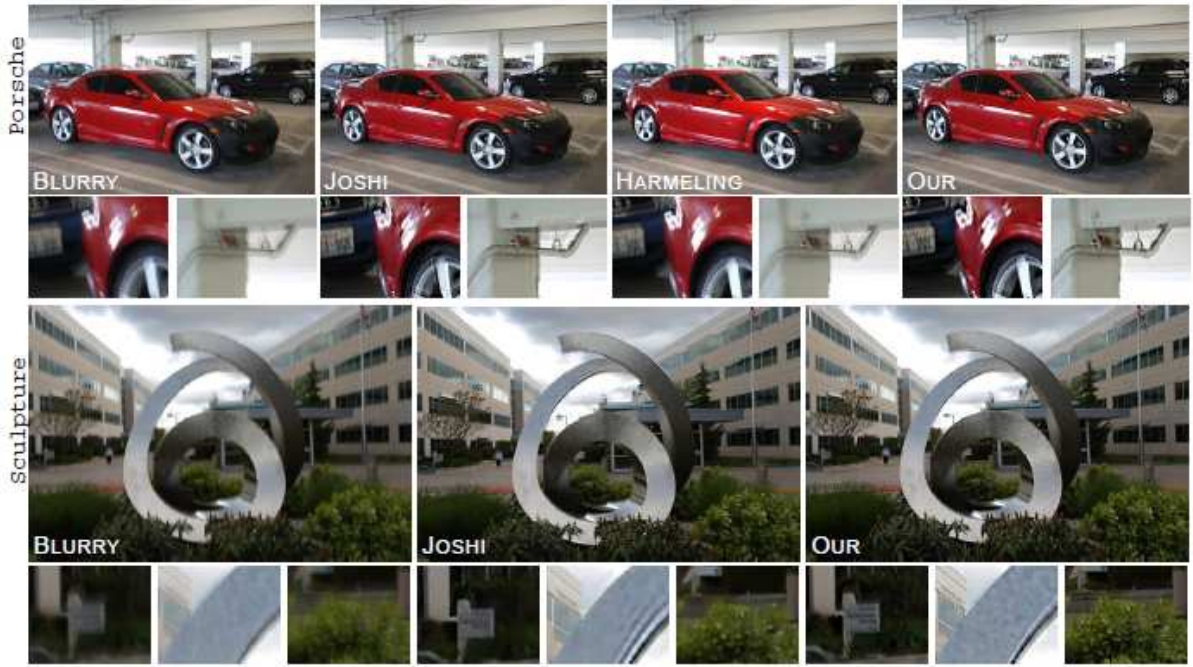}
\caption{Non-uniform deblurring comparisons with  Joshi~\cite{JoshiKZS10} and Harmeling~\cite{HarmelingHS_NIPS10} using the real-world images {\ttfamily Porsche}  and {\ttfamily Sculpture} provided in \cite{JoshiKZS10}.  Note that Harmeling \emph{et al.} do not provide a deblurring result for the {\ttfamily Sculpture} image.}
\label{fig:Joshi}
\end{figure*}

{\bf Comparisons with  Joshi \emph{et al.}~\cite{JoshiKZS10} and Harmeling \emph{et al.}~\cite{HarmelingHS_NIPS10}:} Joshi \emph{et al.}~present a deblurring algorithm that relies upon additional hardware for estimating camera motion~\cite{JoshiKZS10}.  However, even without this additional hardware assistance, our algorithm still produces a better sharp estimate of the {\small\ttfamily Porsche} and {\small\ttfamily Sculpture} images from~\cite{JoshiKZS10}, with fewer ringing artifacts and higher resolution details.  See Figure~\ref{fig:Joshi} for the results, where Harmeling \emph{et al.} have also produced results for the {\small\ttfamily Porsche} image.

\begin{figure*}[t]
\centering
\includegraphics[width=16cm]{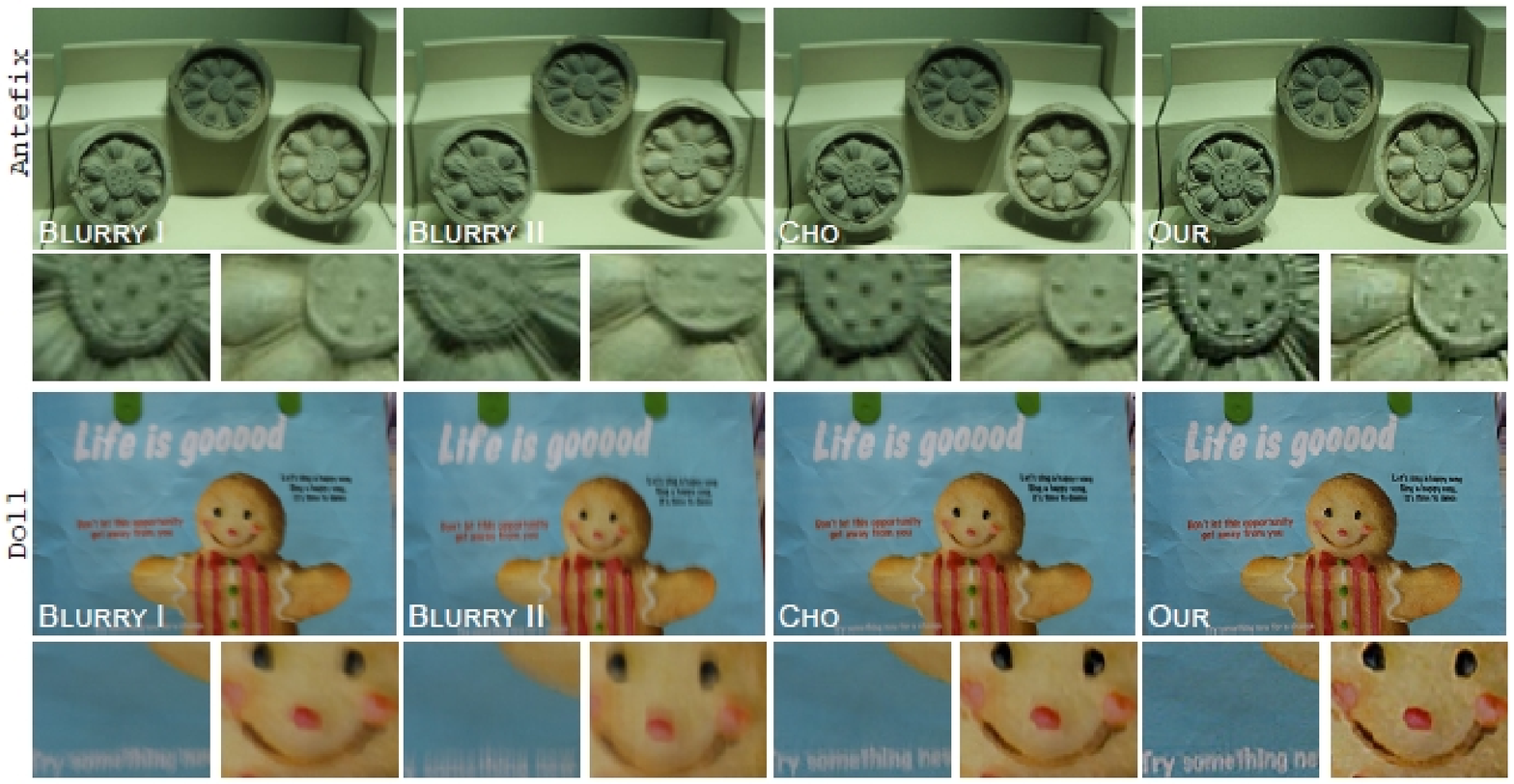}
\caption{Non-uniform deblurring comparisons with Cho \emph{et al.}~\cite{ChoCTL12} on the real-world images {\ttfamily Antefix} and {\ttfamily Doll} from~\cite{ChoCTL12}.  Note that the method of Cho \emph{et al.}~requires two blurry images as input while we ran our algorithm using only the first blurry image in each test pair.}
\label{fig:Cho}
\end{figure*}

{\bf Comparison with  Cho \emph{et al.}~\cite{ChoCTL12}:} Finally, we evaluate deblurring results using the {\small\ttfamily Antefix} and {\small\ttfamily Doll} images from \cite{ChoCTL12}.  The method of Cho \emph{et al.}~requires two blurry images of the same scene as input while we ran our algorithm using only the first blurry image in each test pair. Despite this significant disadvantage, our method still produces higher quality estimates in both cases.  The results are shown in Figure~\ref{fig:Cho}.

\section{Conclusion}
\label{sec:con}

This paper presents a strikingly simple yet effective method for non-uniform camera shake removal based upon a principled, transparent cost function that is open to analysis and further extensions/refinements.  Moreover, both theoretical and extensive empirical evidence are provided demonstrating the efficacy of the adaptive approach to sparse regularization which emerges from our model.  Extending the current framework to handle multiple images and video represents a worthwhile topic for future research.

\appendix

\section{Derivation of the Algorithm} \label{apd:algo_derivation}

Blind deblurring is achieved by minimizing the cost function from (\ref{eq:cost_fun_non_uni}).  This can be accomplished by minimizing a rigorous upper bound $\mathcal{L}(\x, \w, \gam)$ defined as
\begin{eqnarray}\label{eq:cost_fun_upperbound_app}
\mathcal{L}(\x, \w, \gam, \lambda) \triangleq \frac{1}{\lambda} \Vert \y - \HH \x \Vert_2^2 +  \sum_i \left[\frac{x_i^2}{\gamma_i}  + \log(\lambda + \gamma_i \Vert  \wbar_i \Vert^2_2)\right],
\end{eqnarray}
which is obtained by using the fact that
\begin{equation}
g(x_i,\w,\lambda) = \min_{\gamma_i \geq 0}  \frac{x_i^2}{\gamma_i}  + \log(\lambda + \gamma_i \Vert  \wbar_i \Vert^2_2).
\end{equation}
This expression can be shown by optimizing over $\gamma_i$, plugging in the resulting value which can be obtained in closed-form, and then simplifying. $\mathcal{L}(\x, \w, \gam, \lambda)$ can be iteratively minimized by optimizing $\x$, $\w$, $\gam$, and $\lambda$ with similar convergence properties to the EM algorithm.  The resulting procedure is summarized in Algorithm~\ref{algo:algo_non_uniform_BDB}.  We now detail each constituent subproblem.

{\flushleft\bf $\x$-subproblem:}
With other variables fixed, the latent image $\x$  is estimated via weighted least squares giving
\begin{eqnarray}\label{eq:image_update}
{
\begin{split}
\x^{\rm opt} =  \left[\frac{{\HH}^T {\HH}}{{\lambda}} +  {\GM}^{-1} \right]^{-1}  \frac{{\HH}^T \y}{{\lambda}},
 \end{split}
 }
\end{eqnarray}
where   ${\GM} = {\rm diag} [{\boldsymbol{\gamma}}]$.  This can be computed efficiently using EFF and fast Fourier transforms \cite{HarmelingHS_NIPS10}.

{\flushleft\bf $\gam$-subproblem:}
The optimization over each $\gamma_i$ is separable, thus can be solved independently via
\begin{eqnarray}\label{eq:gamma_i_org_problem}
 \begin{split}
\min_{\gamma_i \ge 0}   \left[\frac{x_i^2}{\gamma_i}  + \log \left(\lambda + \gamma_i \Vert \wbar_i \Vert^2_2\right)\right].
 \end{split}
\end{eqnarray}
We can rewrite (\ref{eq:gamma_i_org_problem}) equivalently as
\begin{eqnarray}\label{eq:gamma_i_eqv_problem}
 \begin{split}
\min_{\gamma_i \ge 0}   \left[  \frac{x_i^2}{\gamma_i}   +  \log \gamma_i +  \log \left(\frac{\Vert \wbar_i  \Vert^2_2}{\lambda} + \gamma_i^{-1}\right)\right],
 \end{split}
\end{eqnarray}
where the irrelevant $\log \lambda$ term has been omitted.
As no closed form solution is available for (\ref{eq:gamma_i_eqv_problem}), we instead use principles from convex analysis to form the strict upper bound
\begin{eqnarray}\label{eq:concave_conj_gamma}
  \frac{z_{i}}{\gamma_i} - \phi^*(z_{i}) \ge \log \left(\frac{\Vert \wbar_i  \Vert^2_2}{\lambda} + \gamma_i^{-1}\right), \quad \forall z_{i} \ge 0,
\end{eqnarray}
where $\phi^*(z_{i})$ is the concave conjugate of the concave function $\phi(\alpha)\triangleq \log (\frac{\Vert \wbar_i \Vert^2_2}{\lambda} + \alpha)$.
It can be shown that equality in (\ref{eq:concave_conj_gamma}) is achieved when
\begin{eqnarray}\label{eq:z_opt}
z_{i}^{\rm opt} = \left.\frac{\partial \phi}{\partial \alpha}\right|_{\alpha = \gamma_i^{-1}} = \frac{1}{\frac{\Vert \wbar_i  \Vert^2_2}{\lambda} + \gamma_i^{-1}}, \quad \forall i.
\end{eqnarray}
Substituting (\ref{eq:concave_conj_gamma}) into  (\ref{eq:gamma_i_eqv_problem}), we obtain the revised subproblem
\begin{eqnarray}\label{eq:gamma_i_revised_problem}
 \begin{split}
\min_{\gamma_i \ge 0}  \left[  \frac{x_i^2 + z_{i}}{\gamma_i}   +  \log \gamma_i \right],
 \end{split}
\end{eqnarray}
which admits the closed-form optimal solution
\begin{eqnarray}\label{eq:gam_update}
 \begin{split}
\gamma_i^{\rm opt} = {x_i}^2 +  z_{i}.
 \end{split}
\end{eqnarray}

{\flushleft\bf $\w$-subproblem:}
Isolating $\w$-dependent terms produces the quadratic minimization problem
\begin{eqnarray}\label{eq:kernel_subproblem}
 \begin{split}
\min_{\w\ge 0} \ \frac{1}{{\lambda}}\Vert \y - \D \w\Vert_2^2 +  \sum_i  \log \left(\frac{\Vert \wbar_i  \Vert^2_2}{\lambda} + \gamma_i^{-1}\right).
 \end{split}
\end{eqnarray}
Because again there is no closed-form solution, we resort to similar bounding techniques as used above, incorporating the bound
\begin{eqnarray}\label{eq:concave_conj_k}
\Vert \wbar_i \Vert^2_2 v_i - \psi^*(v_i) \ge \log \left(\frac{\Vert \wbar_i \Vert^2_2}{\lambda} + \gamma_i^{-1}\right), \quad \forall v_i \ge 0,
\end{eqnarray}
where $\psi^*$ is the concave conjugate of the concave function $\psi(\alpha)\triangleq \log(\frac{\alpha}{\lambda} + \gamma_i^{-1})$.
Similar to before, equality is achieved with
\begin{eqnarray}\label{eq:v_opt}
v_i^{\rm opt} = \left. \frac{\partial \psi_i}{\partial \alpha}\right|_{\alpha = \Vert \wbar_i \Vert^2_2} = \frac{z_i}{\lambda}, \quad \forall i.
\end{eqnarray}
Plugging (\ref{eq:concave_conj_k}) into (\ref{eq:kernel_subproblem}), we obtain the minimization problem
\begin{eqnarray}\label{eq:k_subproblem_eqv}
 \begin{split}
&\min_{\w\ge 0}  \frac{1}{{\lambda}}\Vert \y - \D \w\Vert_2^2 +  \sum_i v_i\Vert \wbar_i \Vert^2_2\\
=&\min_{\w\ge 0}   \Vert \y - \D \w\Vert_2^2 +   \w^T \left(\sum_i z_i \B_i^T\B_i \right) \w
 \end{split}
\end{eqnarray}
which can be solved efficiently using standard convex programming techniques.

{\flushleft\bf $\lambda$-subproblem:}
Finally,  the  update rule for the noise level $\lambda$ can be obtained through similar analysis.  Omitting the terms irrelevant to $\lambda$ we must solve
\begin{eqnarray}\label{eq:lambda_sub_problem}
\min_{\lambda\ge 0} \frac{1}{\lambda} \left( \Vert \y-\HH \x \Vert_2^2 + d \right) + n \log \lambda + \sum_i \log \left( \frac{\Vert \wbar_i \Vert^2_2}{\lambda} + \gamma_i^{-1} \right),
\end{eqnarray}
where $n$ is the dimensionality of $\y$ and we have added a small constant $d$ to the quadratic data fit term to prevent it from ever going to exactly zero.  As before there is no closed-form solution, so we invoke the bound
\begin{eqnarray}\label{eq:concave_conj_lambda}
\frac{\beta}{\lambda} - \varphi^*(\beta) \ge \sum_i \log \left( \frac{\Vert \wbar_i \Vert^2_2}{\lambda} + \gamma_i^{-1} \right), \quad \forall \beta\ge 0,
\end{eqnarray}
where $\varphi^*$ is the concave conjugate of $\varphi(\alpha)\triangleq \sum_i \log \left( \alpha \Vert \wbar_i \Vert^2_2 + \gamma_i^{-1} \right)$,
Equality is achieved with
\begin{eqnarray}\label{eq:beta_opt}
\beta^{\rm opt} = \left. \frac{\partial \varphi}{\partial \beta}\right|_{\beta = \lambda^{-1}} = \sum_i \frac{\Vert \wbar_i \Vert^2_2}{\frac{\Vert \wbar_i \Vert^2_2}{\lambda} + \gamma_i^{-1}}.
\end{eqnarray}
Plugging (\ref{eq:concave_conj_lambda}) into (\ref{eq:lambda_sub_problem}), we obtain the problem
\begin{eqnarray}\label{eq:lambdal_subproblem_eqv}
 \begin{split}
\min_{\lambda\ge 0} \frac{1}{\lambda} \left( \Vert \y-\HH \x \Vert_2^2 + d \right)+ n \log \lambda + \frac{\beta}{\lambda} - \phi^*(\beta),
 \end{split}
\end{eqnarray}
leading to the closed-form noise level update
\begin{eqnarray}\label{eq:lambda_opt}
\begin{split}
{\lambda}^{\rm opt} = \frac{\Vert \y - \HH{\x} \Vert_2^2 + \beta + d}{n}.
\end{split}
\end{eqnarray}
Note that $\lambda^{\rm opt}$ has a lower bound of $d/n$.  Thus we may set $d$ so as to reflect some expectation regarding the minimum possible amount of noise or modeling error.  In practice we simple choose $d = n 10^{-4}$ for all experiments.

\section{Proof of Theorem~\ref{thm:spa_adap_sparsity}} \label{apd:proof}
For the first property, it is useful to re-express $h(z; \rho)$ using the equivalent variational form
\begin{eqnarray}\label{eq:h_rho}
h(\nx; \rho) = \min_{\gamma \ge 0} \frac{z^2}{\gamma} + \log(\rho + \gamma), \quad \forall \nx \ge 0,
\end{eqnarray}
which can be verified straightforwardly by calculating the minimizing $\gamma^{\rm opt}$ and plugging it back into (\ref{eq:h_rho}).
As $\psi(\gamma) \triangleq \log (\rho+\gamma)$ is  a concave, non-decreasing function of $\gamma$,   we can always express $\psi(\gamma)$ as
\begin{equation}
\psi(\gamma) = \min_{\nz\ge 0} \nz \gamma - \psi^*(\nz),
\end{equation}
where $ \psi^*(\nz)$ is the concave conjugate \cite{cvx} of $\psi(\gamma)$.  Therefore, it follows that
\begin{equation} \label{eq:gvb_cost_new2}
h(\nx; \rho)  = \min_{\gamma, \nz \ge 0} \frac{\nx^2}{\gamma}  + \nz  \gamma - \psi^*(\nz).
\end{equation}
Optimizing over $\gamma$ for fixed $\nx$ and $\nz$, the optimal solution is
\begin{eqnarray}
\gamma^{\rm opt} = \nz^{-1/2} \nx.
\end{eqnarray}
Plugging this result into (\ref{eq:gvb_cost_new2}) gives
\begin{eqnarray}
\begin{split}
 h(\nx; \rho) & = \min_{\nz \ge 0} \frac{\nx^2}{\nz^{-1/2} \nx} + \nz \nz^{-1/2} \nx  - \psi^*(\nz) = \min_{\nz \ge 0}  2\nz^{1/2} \nx - \psi^*(\nz).
\end{split}
\end{eqnarray}
This implies that $h(\nx; \rho)$ can be expressed as a minimum over upper-bounding hyperplanes in $\nx$, with different $\nz$ implying different slopes.  Any function expressable in this form is necessarily concave, and also non-decreasing since \mbox{$\nz\ge0$} \cite{cvx}.

For the second property, we first define
\begin{eqnarray}
 \hrho{\rho_{\alpha}}(\nz) & \triangleq & h(\sqrt{\nz}; \rho=\rho_{\alpha}) = \min_{\gamma \ge 0} \frac{\nz}{\gamma} + \log(\rho_{\alpha} + \gamma). \label{eq:h_z}
\end{eqnarray}
Using results from convex analysis and conjugate duality, it can be shown that the minimizing $(\gamma^{\rm opt}_{\rho_{\alpha}})^{-1}$  for (\ref{eq:h_z}) represents the gradient of $\hrho{\rho_{\alpha}}(\nz)$ with respect to $\nz$, meaning $\frac{\partial\hrho{\rho_{\alpha}}(\nz)}{ \partial \nz} \equiv (\gamma^{\rm opt}_{\rho_{\alpha}})^{-1}$.
Assuming $\rho_1<\rho_2$,
then the minimizing value of $\gamma_1^{\rm opt}$ and $\gamma_2^{\rm opt}$  associated with $\rho_1$ and $\rho_2$ will always satisfy $\gamma_1^{\rm opt} < \gamma_2^{\rm opt}$, implying $\frac{\partial \hrho{\rho_1}(\nz)}{\partial \nz}>\frac{\partial \hrho{\rho_2}(\nz)}{\partial \nz}$.  This occurs because
\begin{equation}\nonumber
\gamma_1^{\rm opt}  =  \arg\min_{\gamma} \frac{\nz}{\gamma} + \log(\rho_1+\gamma )  =  \arg\min_{\gamma} \frac{\nz}{\gamma} + \log(\rho_2+\gamma) + \log \left( \frac{ \rho_1+\gamma}{\rho_2+\gamma} \right).
\end{equation}
The last term, which is monotonically increasing from $\log\left(\rho_1/\rho_2 \right) < 0$ to zero, implies that there is always an extra monotonically increasing penalty on $\gamma$, when $\rho_1 < \rho_2$.  Since we are dealing with continuous functions here, the minimizing $\gamma$ will therefore necessarily be smaller, thus
 $\frac{\partial \hrho{\rho_1}(\nz)}{\partial \nz}>\frac{\partial \hrho{\rho_2}(\nz)}{\partial \nz}$ at any point $\nz$.
From (\ref{eq:h_z}) and  $\nz \triangleq \nx^2$, we can readily compute the expression for $\frac{\partial h(\nx; \rho)}{\partial \nx}$ as
\begin{eqnarray}\label{eq:dg0}
\begin{split}
\frac{\partial h(\nx; \rho)}{\partial \nx} &= \frac{\partial\hrho{\rho}(\nz)}{\partial \nz} \frac{d\nz}{d\nx} = 2\nx \frac{\partial\hrho{\rho}(\nz)}{\partial \nz}.
\end{split}
\end{eqnarray}
 Given that $\nx \ge 0$ by definition,  we therefore have $\frac{\partial h(\nx; \rho_1)}{\partial \nx} > \frac{\partial h(\nx; \rho_2)}{ \partial\nx}$.

Furthermore, we want to show that $ h(\nx; \rho_1) \prec  h(\nx; \rho_2)$ given $\rho_1<\rho_2$.
For this purpose it is sufficient to show that $\frac{\partial^2 h(\nx; \rho)}{\partial \nx^2}/\frac{\partial h(\nx; \rho)}{\partial \nx}$ is an increasing function of $\rho$, which represents an equivalent condition for relative concavity to one given by Definition~\ref{def:1}~\cite{RC_Palmer}.

From  (\ref{eq:h_z}) and (\ref{eq:dg0}), we can compute the explicit expression for $\frac{\partial h(\nx; \rho)}{\partial \nx}$ as
\begin{eqnarray}\label{eq:dg}
\begin{split}
\frac{\partial h(\nx; \rho)}{\partial \nx} &= 2\nx \frac{\partial\hrho{\rho}(\nz)}{\partial \nz} = \frac{\nx}{\rho} \left(\sqrt{1+\frac{4\rho}{\nx^2}} - 1\right).
\end{split}
\end{eqnarray}
Using (\ref{eq:dg}) it is also straightforward to derive $\frac{\partial^2 h(\nx; \rho)}{\partial \nx^2}$  as
\begin{eqnarray}
\begin{split}
\frac{\partial^2 h(\nx; \rho)}{\partial \nx^2}&  = 2 \frac{\partial\hrho{\rho}(\nz)}{\partial \nz} - \frac{4}{\nx^2\sqrt{1+\frac{4\rho}{\nx^2}}}.
\end{split}
\end{eqnarray}
We must then show that
\begin{eqnarray}
\frac{\partial^2 h(\nx; \rho)/\partial \nx^2}{\partial h(\nx; \rho)/\partial \nx} = \frac{1}{\nx} - \frac{\frac{4}{\nx^2\sqrt{1+\frac{4\rho}{\nx^2}}}}{\frac{\nx}{\rho} \left(\sqrt{1+\frac{4\rho}{\nx^2}} - 1\right)}
\end{eqnarray}
is an  increasing function of $\rho$.  By neglecting  irrelevant additive and multiplicative factors (and recall that $\nx\ge 0$ from the definition of $h(\nx; \rho)$), this is equivalent to showing that
\begin{eqnarray}
\xi(\rho) = \frac{1}{\rho} \left(\sqrt{1+\frac{4\rho}{\nx^2}} -1 \right)
\end{eqnarray}
is a  decreasing function of $\rho$.  It is easy to check that
\begin{eqnarray}
\begin{split}
\xi'(\rho)
& = \frac{\sqrt{1+\frac{4\rho}{\nx^2}} - 1 - \frac{2\rho}{\nx^2}}{\sqrt{1+\frac{4\rho}{\nx^2}}} <0.
\end{split}
\end{eqnarray}
Therefore, $\xi(\rho)$ is a decreasing function of $\rho$, implying that $\frac{\partial^2 h(\nx; \rho)}{\partial \nx^2}/\frac{\partial h(\nx; \rho)}{\partial \nx}$ is an increasing function of $\rho$,
 completing the proof.~\myendofproof

\bibliographystyle{IEEEbib}
\bibliography{NU_BDB}

\begin{thebibliography}{10}

\bibitem{Fergus06removingcamera}
Rob Fergus, Barun Singh, Aaron Hertzmann, Sam~T. Roweis, and William~T.
  Freeman,
\newblock ``Removing camera shake from a single photograph,''
\newblock in {\em SIGGRAPH}, 2006.

\bibitem{hqdeblurring_siggraph2008}
Qi~Shan, Jiaya Jia, and Aseem Agarwala,
\newblock ``High-quality motion deblurring from a single image,''
\newblock in {\em SIGGRAPH}, 2008.

\bibitem{LevinWDF11_PAMI}
Anat Levin, Yair Weiss, Fr{\'e}do Durand, and William~T. Freeman,
\newblock ``Understanding blind deconvolution algorithms,''
\newblock {\em IEEE Trans. Pattern Anal. Mach. Intell.}, vol. 33, no. 12, pp.
  2354--2367, 2011.

\bibitem{fast_motion_deblur_2009}
Sunghyun Cho and Seungyong Lee,
\newblock ``Fast motion deblurring,''
\newblock in {\em SIGGRAPH ASIA}, 2009.

\bibitem{XuJ10_ECCV}
Li~Xu and Jiaya Jia,
\newblock ``Two-phase kernel estimation for robust motion deblurring,''
\newblock in {\em ECCV}, 2010.

\bibitem{norm_sparse}
Dilip Krishnan, Terence Tay, and Rob Fergus,
\newblock ``Blind deconvolution using a normalized sparsity measure,''
\newblock in {\em CVPR}, 2011.

\bibitem{WangYYZ08}
Yilun Wang, Junfeng Yang, Wotao Yin, and Yin Zhang,
\newblock ``A new alternating minimization algorithm for total variation image
  reconstruction,''
\newblock {\em SIAM J. Imaging Sciences}, vol. 1, no. 3, pp. 248--272, 2008.

\bibitem{BeckT09}
Amir Beck and Marc Teboulle,
\newblock ``A fast iterative shrinkage-thresholding algorithm for linear
  inverse problems,''
\newblock {\em SIAM J. Imaging Sciences}, vol. 2, no. 1, pp. 183--202, 2009.

\bibitem{Whyte_non-uniformdeblurring}
Oliver Whyte, Josef Sivic, Andrew Zisserman, and Jean Ponce,
\newblock ``Non-uniform deblurring for shaken images,''
\newblock in {\em CVPR}, 2010.

\bibitem{Gupta10singleimage}
Ankit Gupta, Neel Joshi, C.~Lawrence Zitnick, Michael Cohen, and Brian Curless,
\newblock ``Single image deblurring using motion density functions,''
\newblock in {\em ECCV}, 2010.

\bibitem{HarmelingHS_NIPS10}
Stefan Harmeling, Michael Hirsch, and Bernhard Sch{\"o}lkopf,
\newblock ``Space-variant single-image blind deconvolution for removing camera
  shake,''
\newblock in {\em NIPS}, 2010.

\bibitem{HirschSHS_ICCV11}
Michael Hirsch, Christian~J. Schuler, Stefan Harmeling, and Bernhard
  Sch{\"o}lkopf,
\newblock ``Fast removal of non-uniform camera shake,''
\newblock in {\em ICCV}, 2011.

\bibitem{hu_bmvc2012}
Zhe Hu and Ming-Hsuan Yang,
\newblock ``Fast non-uniform deblurring using constrained camera pose
  subspace,''
\newblock in {\em BMVC}, 2012.

\bibitem{ChoCTL12}
Sunghyun Cho, Hojin Cho, Yu-Wing Tai, and Seungyong Lee,
\newblock ``Registration based non-uniform motion deblurring,''
\newblock {\em Comput. Graph. Forum}, vol. 31, no. 7-2, pp. 2183--2192, 2012.

\bibitem{Xu_depth-awaremotion}
Li~Xu and Jiaya Jia,
\newblock ``Depth-aware motion deblurring,''
\newblock in {\em ICCP}, 2012.

\bibitem{non_uniform_restoration_chp3}
Michal Sorel and Filip Sroubek,
\newblock {\em Image Restoration: Fundamentals and Advances},
\newblock CRC Press, 2012.

\bibitem{JiHui12}
Hui Ji and Kang Wang,
\newblock ``A two-stage approach to blind spatially-varying motion
  deblurring,''
\newblock in {\em CVPR}, 2012.

\bibitem{Tai09kasittr}
Yu-Wing Tai, Ping Tan, and Michael~S. Brown,
\newblock ``{Richardson-Lucy} deblurring for scenes under a projective motion
  path,''
\newblock {\em IEEE Trans. Pattern Anal. Mach. Intell.}, vol. 33, no. 8, pp.
  1603--1618, 2011.

\bibitem{JoshiKZS10}
Neel Joshi, Sing~Bing Kang, C.~Lawrence Zitnick, and Richard Szeliski,
\newblock ``Image deblurring using inertial measurement sensors,''
\newblock in {\em ACM SIGGRAPH}, 2010.

\bibitem{Filter_Flow}
Steven~M. Seitz and Simon Baker,
\newblock ``Filter flow,''
\newblock in {\em ICCV}, 2009.

\bibitem{HirschSSH10}
Michael Hirsch, Suvrit Sra, Bernhard Sch{\"o}lkopf, and Stefan Harmeling,
\newblock ``Efficient filter flow for space-variant multiframe blind
  deconvolution,''
\newblock in {\em CVPR}, 2010.

\bibitem{Leary}
James~G. Nagy and Dianne~P. O'Leary,
\newblock ``Restoring images degraded by spatially variant blur,''
\newblock {\em SIAM J. Sci. Comput.}, vol. 19, no. 4, pp. 1063--1082, 1998.

\bibitem{LevinWDF11}
Anat Levin, Yair Weiss, Fr{\'e}do Durand, and William~T. Freeman,
\newblock ``Efficient marginal likelihood optimization in blind
  deconvolution,''
\newblock in {\em CVPR}, 2011.

\bibitem{Wipf_VEM_NIPS05}
J.~A. Palmer, D.~P. Wipf, K.~Kreutz-Delgado, and B.~D. Rao,
\newblock ``Variational {EM} algorithms for non-{Gaussian} latent variable
  models,''
\newblock in {\em NIPS}, 2006.

\bibitem{Wipf_Latent_Variable_TIT11}
D.~P. Wipf, B.~D. Rao, and S.~S. Nagarajan,
\newblock ``Latent variable {Bayesian} models for promoting sparsity,''
\newblock {\em IEEE Trans. Information Theory}, vol. 57, no. 9, pp. 6236--6255,
  2011.

\bibitem{CCCP_NIPS03}
Alan~L. Yuille and Anand Rangarajan,
\newblock ``The concave-convex procedure ({CCCP}),''
\newblock in {\em NIPS}, 2001, pp. 1033--1040.

\bibitem{RC_Palmer}
J.~A. Palmer,
\newblock ``Relatve convexity,''
\newblock Technical report, UCSD, 2003.

\bibitem{RaoECPK03}
Bhaskar~D. Rao, Kjersti Engan, Shane~F. Cotter, Jason~A. Palmer, and Kenneth
  Kreutz-Delgado,
\newblock ``Subset selection in noise based on diversity measure
  minimization,''
\newblock {\em IEEE Trans. Signal Processing}, vol. 51, no. 3, pp. 760--770,
  2003.

\bibitem{sps_deblur}
A.~Levin, R.~Fergus, F.~Durand, and W.~T. Freeman,
\newblock ``Deconvolution using natural image priors,''
\newblock Tech. {R}ep., MIT, 2007.

\bibitem{KrishnanF09_NIPS}
Dilip Krishnan and Rob Fergus,
\newblock ``Fast image deconvolution using hyper-{Laplacian} priors,''
\newblock in {\em NIPS}, 2009.

\bibitem{Cho_PAMI}
Taeg~Sang Cho, C.~Lawrence Zitnick, Neel Joshi, Sing~Bing Kang, Rick Szeliski,
  and William~T. Freeman,
\newblock ``Image restoration by matching gradient distributions,''
\newblock {\em IEEE Trans. Pattern Anal. Mach. Intell.}, vol. 34, no. 4, pp.
  683--694, 2012.

\bibitem{cvx}
Stephen Boyd and Lieven Vandenberghe,
\newblock {\em Convex Optimization},
\newblock Cambridge University Press, Cambridge, UK, 2004.

\end{thebibliography}

\end{document}